%% file: main.tex
\documentclass{article}

\usepackage[final,nonatbib]{neurips_2022}

\usepackage[utf8]{inputenc} 
\usepackage[T1]{fontenc}    
\usepackage{hyperref}       
\hypersetup{colorlinks}
\usepackage{url}            
\usepackage{booktabs}       
\usepackage{amsfonts}       
\usepackage{nicefrac}       
\usepackage{microtype}      
\usepackage{xcolor}         
\usepackage{amsmath}
\usepackage{enumitem}
\usepackage{graphicx}
\usepackage{subcaption}
\usepackage{wrapfig}
\usepackage{multirow}
\usepackage{array}
\usepackage{cite}
\setlist[itemize]{leftmargin=*}

\newcommand{\pub}[1]{{\color{gray}{\tiny{[{#1}]}}}}

\title{Segment and Track Anything}

\author{%
  Yangming Cheng,  Liulei Li,  Yuanyou Xu,  Xiaodi Li,\\
  \textbf{Zongxin Yang$^{\star}$,  Wenguan Wang$^{ \star}$,  Yi Yang$^{ \dag}$}
  \\
  ReLER, CCAI, Zhejiang University\\
  {\small\texttt{\{chengyangming, liliulei, yoxu\}@zju.edu.cn, lixiaodi4work@gmail.com}} \\
  {\small\texttt{\{yangzongxin, wenguanwang, yangyics\}@zju.edu.cn}}\\
}

\begin{document}

\maketitle

\let\thefootnote\relax\footnote{$\star$: the project leader.}
\let\thefootnote\relax\footnote{$\dag$: the corresponding author.}

\input{Sections/abstract}

\input{Sections/introduction}

\input{Sections/preliminaries}

\input{Sections/methodology}

\input{Sections/experiments}

\input{Sections/applications}

\input{Sections/conclusion}


{
\bibliographystyle{splncs04}
\bibliography{reference}
}

\end{document}

%% file: Sections/abstract.tex
\begin{abstract}
This report presents a framework called Segment And Track Anything (SAM-Track) that allows users to precisely and effectively segment and track any object in a video. Additionally, SAM-Track employs multimodal interaction methods that enable users to select multiple objects in videos for tracking, corresponding to their specific requirements. These interaction methods comprise click, stroke, and text, each possessing unique benefits and capable of being employed in combination. As a result, SAM-Track can be used across an array of fields, ranging from drone technology, autonomous driving, medical imaging, augmented reality, to biological analysis. SAM-Track amalgamates Segment Anything Model (SAM), an interactive key-frame segmentation model, with our proposed AOT-based tracking model (DeAOT), which secured 1st place in four tracks of the VOT 2022 challenge, to facilitate object tracking in video. In addition, SAM-Track incorporates Grounding-DINO, which enables the framework to support text-based interaction. We have demonstrated the remarkable capabilities of SAM-Track on DAVIS-2016 Val (92.0\%), DAVIS-2017 Test (79.2\%) and its practicability in diverse applications. The project page is available at: \url{https://github.com/z-x-yang/Segment-and-Track-Anything}.

\end{abstract}

%% file: Sections/introduction.tex
\section{Introduction}\label{sec:introduction}

As a fundamental and challenging task in computer vision, video segmentation has great potential in a large range of real-world applications, including drone industry, autonomous driving, medical image processing, augmented reality, and biological analysis, to name a few~\cite{zhou2022survey}. Due to diverse demands from different fields, video segmentation models are required to support multiple interaction modes. Therefore, it is divided into several subtasks, \emph{i.e}, unsupervised (automatic) video segmentation~\cite{wang2017saliency,lu2019see,wang2019zero,tom,wang2019learning}, semi-supervised (mask-guided semi-automatic) video segmentation~\cite{cfbi, cfbip, deaot, aot, aost,lu2020video}, interactive (scribble or click based) video segmentation~\cite{mivos}, and language-induced video segmentation~\cite{locater}. Each subtask has its corresponding segmentation mode and focuses on specific fields. However, the development of a unified framework for video segmentation that effectively meets the distinct needs of each domain has not been fully explored.\par

Recently, the Segment Anything Model (SAM)~\cite{sam} has gained significant attention as a large-scale model in the field of computer vision. SAM is capable of producing high-quality object masks from flexible prompts, including points, boxes, and text, which greatly enhances its user-friendliness. In addition, SAM has demonstrated strong zero-shot performance on a series of segmentation tasks, further expanding its applicability. \par

Despite the above advantages, applying image-based SAM directly to video segmentation produces suboptimal results since the temporal coherence among frames is not taken into account. Besides, SAM does not output semantic labels and the textual prompt is not efficient enough to support referring object segmentation and other high level tasks which demand semantic-level understanding. As a result, SAM is not suitable for video segmentation tasks that highlights the correlation of objects across frames, in particular language guided ones, thus limiting the utility of SAM in applications such as AR and autonomous driving.\par

Considering the strengths and weaknesses of SAM, we adapt SAM to the field of video segmentation with significant modifications. We introduce SAM-Track, a unified video framework that allows users to perform object tracking and segmentation in videos through multimodal interaction methods or automatic methods. To be specific, SAM-Track uses SAM to interactively obtain segments of keyframes as the reference for DeAOT~\cite{deaot},a highly efficient multi-object tracking mode that provides tracking speeds for multiple objects comparable to those of other VOS models used for tracking a single object. DeAOT then propagates the reference frames to track multiple objects in subsequent frames of the video. In order to enhance the language understanding capability of SAMTrack, we integrate Grounding-DINO~\cite{grounding-dino} into the system. With the powerful open-set object detection ability of Grounding-DINO, SAM-Track can interactively select objects in videos for tracking and segmenting through natural language.\par

In conclusion, SAM-Track has remarkable tracking and segmentation abilities and two user-friendly tracking modes to adapt to different requirements for diverse applications. For interactive mode, SAM-Track can track and segment any object in videos using multimodal interaction methods such as clicking, drawing, and text input. These interactive methods provide users with flexible options to select objects of interest in the first frame of videos. Nevertheless, Automatic mode allows SAM-Track track any new objects appearing in any frame of the video. This feature enables SAM-Track to be applied to AR and autonomous driving.

We conduct extensive experiments on two popular multi-object benchmarks for VOS, \emph{i.e}., DAVIS 2016~\cite{davis2016} and DAVIS 2017~\cite{davis2017}, to validate the effectiveness and efficiency of the proposed SAM-Track. To further explore the usability and extensibility of our proposed approach, we also conduct application based experiments in a number of areas. The results additionally demonstrate the great promise of our SAM-Track for real-world scenarios.\par

Overall, our contributions are summarized below:
\begin{itemize}
\vspace{-0.5em}
\item  Instead of separately processing each frame with SAM, we propose a unified video segmentation framework SAM-Track, applying and further extending SAM to video segmentation. SAM-Track enables users to accurately and efficiently track and segment anything in the video under the consideration of temporal coherence.
\vspace{-0.5em}
\item By combining SAM, DeAOT, and Grounding-DINO, SAM-Track supports two tracking modes: interactive mode for user-friendly multimodal selection of objects that will be tracked throughout the video, and automatic mode for automatically track any new objects that appear in subsequent frames of the video.

\vspace{-0.5em}
\item  The promising results demonstrate the outstanding performance of our proposed method and its great potential for real-world applications.
\end{itemize}

%% file: Sections/preliminaries.tex
\section{Preliminaries}\label{sec:preliminaries}

 \textbf{DeAOT.} DeAOT is an AOT-based~\cite{aost, aot, yang2021towards, paot} VOS model that employs an identification mechanism to associate multiple targets in the same high-dimensional embedding space, enabling it to track multiple objects at the same speed as tracking a single object. Additionally, DeAOT uses hierarchical Gated Propagation Module (GPM) to separately propagate both object-agnostic and object-specific embeddings from past frames to the current frame to preserve the object-agnostic visual information in deep propagation layers. As a semi-supervised video segmentation model, DeAOT has shown superior performance and placed first in four competitions of the VOT2022 challenge~\cite{vot2022}. \par

\textbf{Segment Anything Model (SAM).} Recently, SAM has garnered considerable attention as a large-scale model in the field of image segmentation. Through specially designed training methods and a large-scale training data SA-1B~\cite{sam}, it offers not only support for interactive segmentation methods but also delivers outstanding zero-shot performance on a wide range of segmentation tasks. These two crucial features significantly enhance the applicability of SAM and make it a promising solution for various computer vision applications.\par

\textbf{Grounding-DINO.} Grounding-DINO is an open-set object detector that integrates language into closed-set detectors~\cite{dino} at multiple stages. It has a good understanding of language and is able to accomplish the Referring Object Detection task. Given text categories or a detailed reference of the target object, Grounding-DINO can detect the target objects and return the minimum external rectangle for each target.\par

%% file: Sections/methodology.tex
\section{Methodology}\label{sec:methodology}
\input{Figures/overview}
In this section, we will present SAM-Track, our proposed unified video segmentation framework that can cater to diverse requirements in various field applications. An overview of SAM-Track is shown in Figure ~\ref{fig:overview}. We will introduce the Interactive Tracking mode in Section~\ref{subsec:3.1} and the Automatic Tracking mode in Section~\ref{subsec:3.2}. Finally, we will present the Fusion Tracking mode in Section~\ref{subsec:3.3}, which can utilize both Interactive and Automatic Tracking modes simultaneously and selectively.

\subsection{Interactive tracking mode}\label{subsec:3.1}
This subsection will introduce the pipeline for the interactive tracking mode of SAM-Track, as well as the details of how it integrates with Grounding-DINO, SAM, and DeAOT for object detection, annotation, and tracking.\\
DeAOT achieves promising results on various benchmarks. However, as a semi-supervised video segmentation model, DeAOT requires reference frame annotations for initialization, which limits its application. \\
An innovative approach to obtain annotations is by utilizing SAM for interactive object segmentation in the reference frame. Leveraging the highly interactive segmentation approach of SAM, annotations can be obtained with accuracy and efficiency for any video. Specifically, we use the interactive methods of SAM - click and box - to segment objects of interest in the reference frame. DeAOT then takes the segmentation result as annotation, using the Gated Propagation Module (GPM) to hierarchically propagate visual embeddings and ID embeddings from past frames to the current frame for tracking objects frame-by-frame. \\
While SAM is a powerful foundation model for interactive object segmentation, it has certain limitations. Specifically, it does not provide sufficient semantic information, and the textual prompt may not efficiently support tasks that demand a more nuanced understanding of object segmentation, such as those involving semantic-level understanding.\\
To alleviate this limitation, we have integrated Grounding-DINO into SAM-Track as a listener. This integration has enabled SAM-Track to utilize the powerful open-set object detection capabilities of Grounding-DINO, which allows for interactive selection of objects in reference frame using natural language commands. In particular, Grounding-DINO takes as input either text categories or detailed descriptions of distinct objects, and outputs the minimum external rectangle for each target. SAM subsequently utilizes these rectangles as box prompts to predict the mask for each object. The resulting masks of objects are then used by DeAOT to track the objects in the video. After combining Grounding-DINO, SAM, and DeAOT, SAM-Track supports multimodal interaction methods. 

\subsection{Automatic tracking mode}\label{subsec:3.2}

This subsection will introduce the automatic tracking mode of SAM-Track, including how it tracks new objects appearing in video and the definition of new objects.\\
The interactive tracking mode of SAM-Track introduced in Section~\ref{subsec:3.1} has great interaction capability, allowing for efficient annotation of objects in reference frames. This mode meets the requirements of most application scenarios. Nevertheless, the interactive tracking mode cannot handle new objects of interest that appear in the video due to the absence of annotations. In order to track new objects appearing in the video, we propose two methods: \textbf{Segment Everything} and \textbf{Object of Interest Segmentation}, to get the annotations for new objects in every nth frame.\\
In the Segment Everything method, we use the segment-everything function of SAM to obtain object masks for every object in the key reference frame. Then DeAOT tracks emerging objects and original objects based on the merged annotations. In contrast, in the Object of Interest Segmentation method, we leverage Grounding-DINO and SAM to obtain annotations for new objects that appear in the video. In detail, Grounding-DINO will detect objects in every nth frame according to predetermined text prompts, \emph{i.e}, ``person''  and ``car''. Then, SAM and DeAOT will handle the annotation and tracking of new objects, as described in Section~\ref{subsec:3.1}\\
Defining new objects is a tough question for automated tracking modes because each object has a unique ID that DeAOT uses to distinguish between objects. If we directly assign new ID to all objects that annotationed in key-frames, it will lead to ID exchange in tracking process (unsuccessful tracking). We use Comparing Mask Results (CMR) to determine new objects. In CMR, we compare the tracking results from DeAOT and the annotation results from SAM in every key-frame, and select objects from the SAM annotations that are not being tracked by DeAOT. \\
Specifically, we define \emph{$N \in \mathbb{R}^{H \times W}$}, \emph{$T_{0} \in \mathbb{R}^{H \times W}$}, \emph{$S \in \mathbb{R}^{H \times W}$} and \emph{t} as new objects mask, the background of DeAOT tracking result, SAM annotation result and minimum threshold for new objects respectively, where \emph{H, W} denote the height and width dimensions. The new objects mask is obtained by using Eq~\ref{equ:1}. 
\begin{equation}\label{equ:1}
\begin{aligned}
N = T_{0} * S  
\end{aligned}
\end{equation}
For an object $x$ that appears in $N$, we define its size in $S$ and $N$ as $x_{s}$ and $x_{n}$ respectively, and if the ratio between $x_{s}$ and $x_{n}$ is bigger than $t$, this object will be defined in new object, formulate by Eq~\ref{equ:2}. Through the use of CMR, we can significantly mitigate the issue of newly detected objects affecting the ID tracking of objects that have already been tracked. 
\begin{equation}\label{equ:2}
\begin{aligned}
CMR(x) = \begin{cases}
1 , \quad if \quad \frac{x_{n}}{x_{s}} > t \\
0 , \quad else 
\end{cases}
\end{aligned}
\end{equation}

\subsection{Fusion tracking mode}\label{subsec:3.3}

SAM-Track is a unified video segmentation framework that supports various combinations of tracking methods. Specifically, each of the interactive methods in SAM-Track can be combined with automatic tracking mode. The Interactive Tracking mode obtains annotations for the first frame of the video, while the Automatic Tracking mode tracks new objects appearing in the video that were not selected in the first frame. The diverse combinations of tracking methods expand the range of applications for SAM-Track.

%% file: Figures/overview.tex
\begin{figure}[t!]
\begin{center}
\vspace{-6.5mm}
\includegraphics[width=1\textwidth]{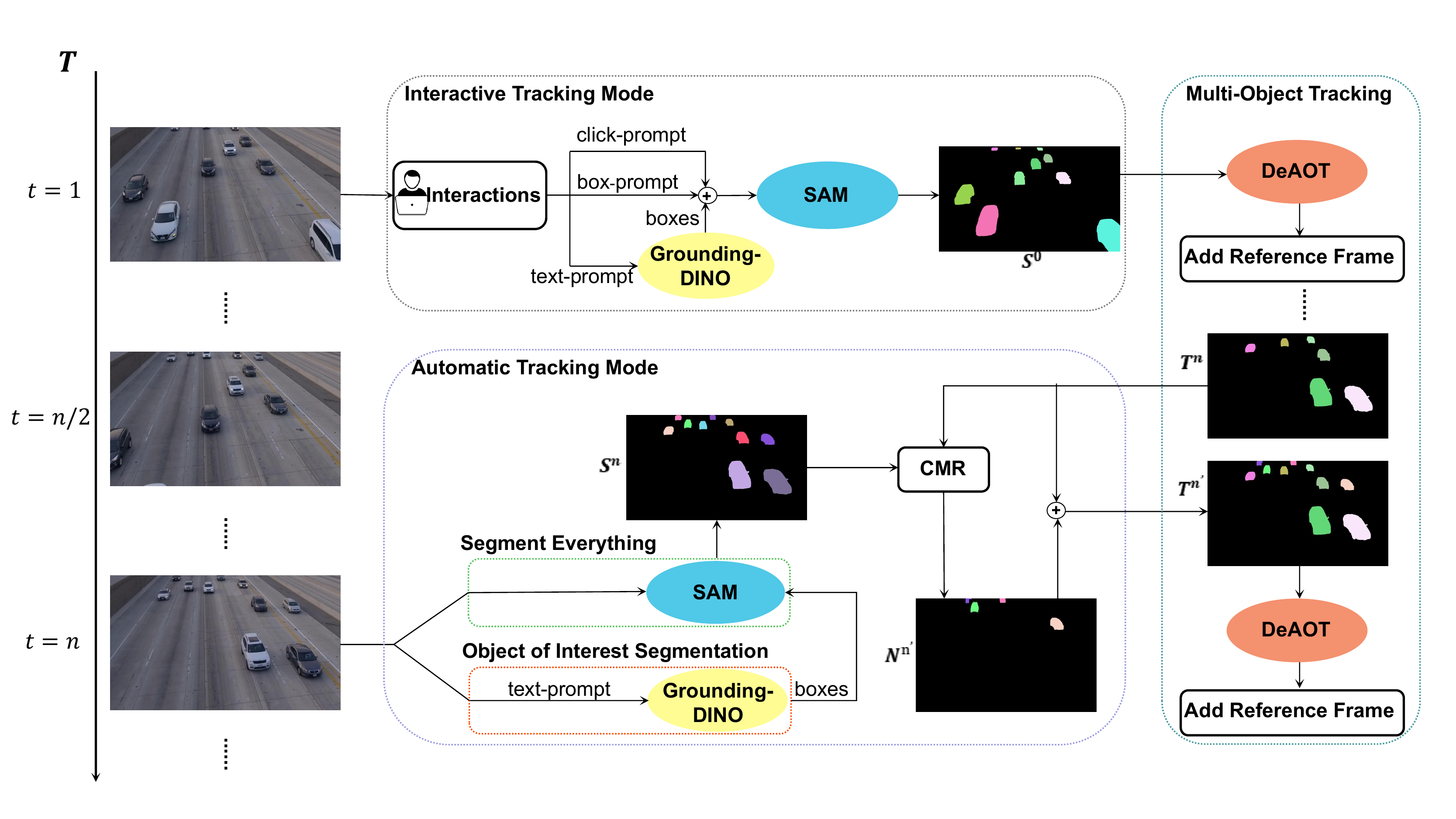}
\caption{The Pipeline of SAM-Track. The Interactive Tracking Mode is used only in the first frame of the video to obtain annotations, while the Automatic Tracking Mode is called every nth frame thereafter. The $S^t$, $T^t$, ${T^n}^{'}$, ${N^n}^{'}$ denote the SAM annotation result, DeAOT tracking result, Refined DeAOT tracking result and Refined new objects mask respectively. It should be noted that if n is greater than the number of video frames, the Automatic Tracking Mode will not be called during the tracking process.}\label{fig:overview}
\end{center}
\end{figure}

%% file: Sections/experiments.tex
\section{Experiments}\label{sec:experiments}
\input{Tables/davis16_17}
\input{Figures/davis_vis}

\subsection{Quantitative Results}

To demonstrate the outstanding performance of our proposed SAM-Track, We conduct experiments on two popular VOS benchmarks (DAVIS-2016 Val and DAVIS-2017 Test). The results are presented in Table~\ref{tab:davis16_17}. It is worth noting that the reference frame annotation was obtained through mouse clicking, an interactive annotation method of SAM-Track. This highlights the efficacy of interactive annotation generation and the outstanding robustness of SAM-Track.

\subsection{Qualitative Results}

We also present some qualitative results in Figure ~\ref{fig:davis_vis}, which demonstrate that SAM-Track is capable of effectively tracking multiple objects simultaneously, even in complex tracking scenarios.

%% file: Tables/davis16_17.tex
\begin{table}[t!]
\begin{center}

\scriptsize
\setlength{\tabcolsep}{4mm}
\begin{tabular}{l| c | c c c| c c c }
\toprule[1.5pt]
    & &  \multicolumn{3}{c}{DAVIS-2016-Val~\cite{davis2016}} & \multicolumn{3}{c}{DAVIS-2017-Test~\cite{davis2017}} \\
 Method & Initialization & Avg & $\mathcal{J}$ & $\mathcal{F}$ & Avg & $\mathcal{J}$ & $\mathcal{F}$ \\
\midrule[1pt]
CFBI\pub{ECCV20}~\cite{cfbi}  & Mask & 89.4  & 88.3 & 90.5 & 75.6 & 71.6 & 79.6 \\
CFBI+\pub{TPAMI21}~\cite{cfbip}  & Mask & 89.9  & 88.7 & 91.1 & 78.0 & 74.4 & 81.6 \\
MiVOS\pub{CVPR21}~\cite{mivos} & Scribble & 91.0 &  89.6 & 92.4 & 78.6 & 74.9 & 82.2 \\
STCN\pub{NeurIPS21}~\cite{cheng2021stcn} & Mask &  91.6 & {90.8}  & {92.5} & 76.1 & 72.7 & 79.6 \\
R50-AOT-L\pub{NeurIPS21}~\cite{aot} & Mask &  {91.1}  & {90.1} & {92.1} & 79.6 & 75.9 & 83.3 \\
XMem\pub{ECCV22}~\cite{xmem} & Mask & 92.0  & 90.7 & 93.2 & 81.2 & 77.6 & 84.7 \\
R50-DeAOT-L\pub{NeurIPS22}~\cite{deaot} & Mask & 92.3 & 90.5 & 94.0 & 80.7 & 76.9 & 84.5\\
SwinB-DeAOT-L\pub{NeurIPS22}~\cite{deaot} & Mask & 92.9 & 91.1 & 94.7 & 82.8 & 78.9 & 86.7\\  
\hline
\hline

\textbf{SAM-Track(Ours)} & Click & 92.0 & 90.3 & 93.6 & 79.2 & 75.3 & 83.1 \\
\bottomrule[1.5pt]
\end{tabular}
\vspace{2mm}
\caption{The quantitative results on the single-object benchmark, DAVIS-2016-Val~\cite{davis2016} and on multi-object benchmarks, DAVIS-2017-Test~\cite{davis2017}. It is worth noting that SAM-Track utilizes the R50-DeAOT-L model for tracking.} \label{tab:davis16_17}
\end{center}
\end{table}

%% file: Figures/davis_vis.tex
\begin{figure}[t!]
\begin{center}
\vspace{-5.5mm}
\includegraphics[width=1\textwidth]{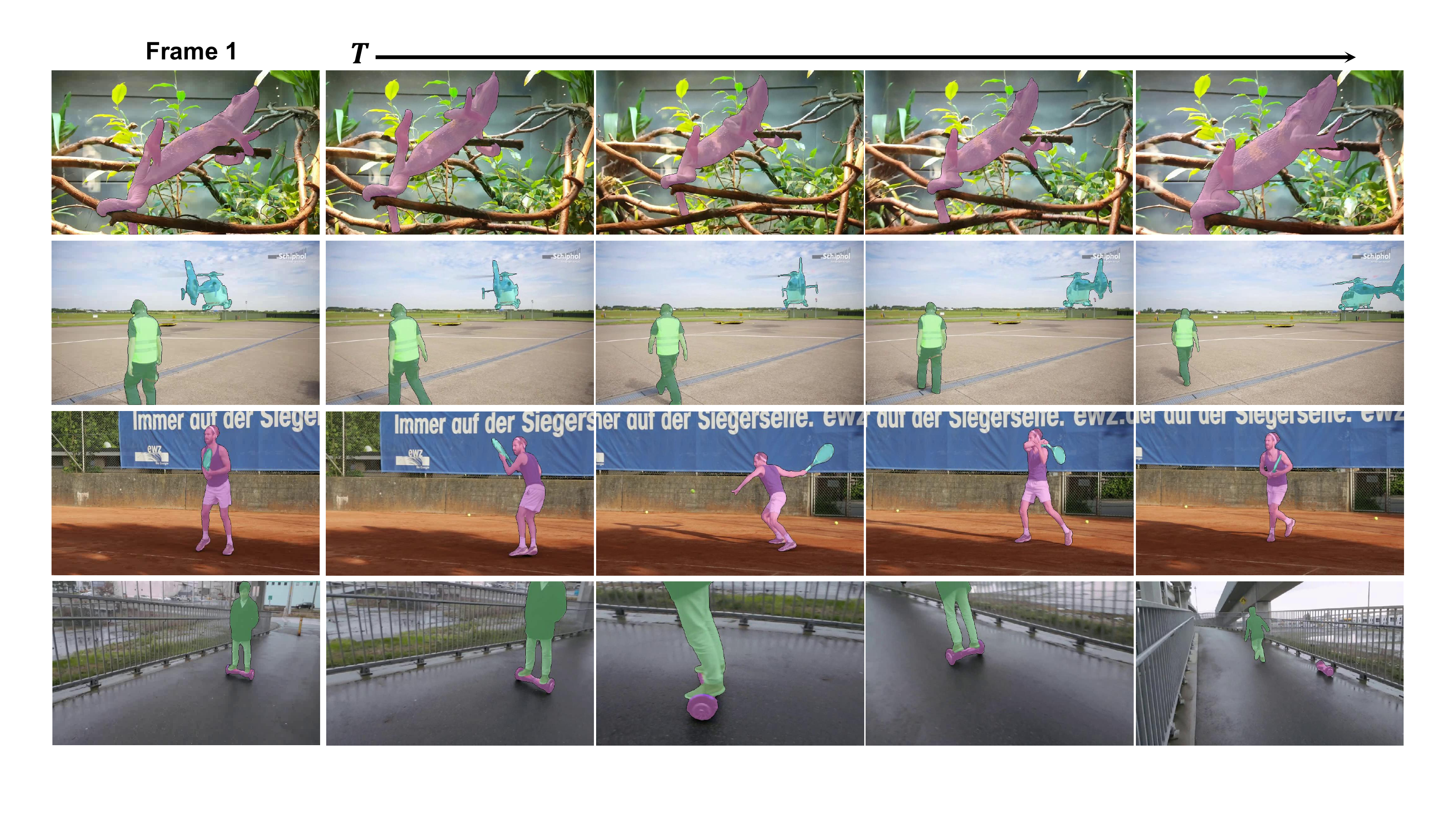}
\caption{Qualitative results of SAM-Track on video from  DAVIS-2016 Val~\cite{davis2016} and DAVIS-2017 Test~\cite{davis2017}.}\label{fig:davis_vis}
\end{center}
\end{figure}

%% file: Sections/applications.tex
\section{Applications}\label{sec:applications}
\input{Figures/app}
SAM-Track, as a unified video segmentation framework, provides two tracking modes and various interactions to easily meet the requirements of different fields. Moreover, the use of DeAOT as a tracking mode in SAM-Track enables it to achieve promising results in complex application scenarios.

\textbf{Sports Analysis.} The automatic tracking mode of SAM-Track can be used for sports analysis. The text prompt ``soccer players'' is set in advance. Additionally, based on the text prompt, SAM-Track can also obtain annotations of the football field by clicking. SAM-Track then continuously tracks soccer players and the football field in the video.

\textbf{Medical Field.} In the medical field, where there is a scarcity of samples for many cells and organs, it is difficult to train a specific tracker for tracking these rare objects. However, SAM-Track's ability to easily track zero-shot objects by clicking makes it very useful. This allows SAM-Track to handle the tracking of these objects without the need for specialized training.

\textbf{Smart City.} The application of VOS models in the field of smart cities is challenging due to the constant appearance of new vehicles in video. Traditional VOS models may not be effective in handling this scenario. However, SAM-Track offers an automatic tracking mode that fully meets these requirements.

\textbf{Autonomous Driving.} The requirements for autonomous driving are similar to those of smart cities, but they also include the need to track pedestrians, pets, and other precise objects. This requires a VOS model that can track a large number of objects simultaneously. SAM-Track has the capability to track numerous objects, making it a suitable option for application in the field of autonomous driving.

To demonstrate the wide range of applications of SAM-Track, we tested it performance in various field. Figure~\ref{fig:app} shows some representative results, indicating that SAM-Track can be effectively applied to diverse fields and produce excellent results using different tracking modes and interaction methods.

%% file: Figures/app.tex
\begin{figure}[t!]
\begin{center}
\vspace{-6.5mm}
\includegraphics[width=1\textwidth]{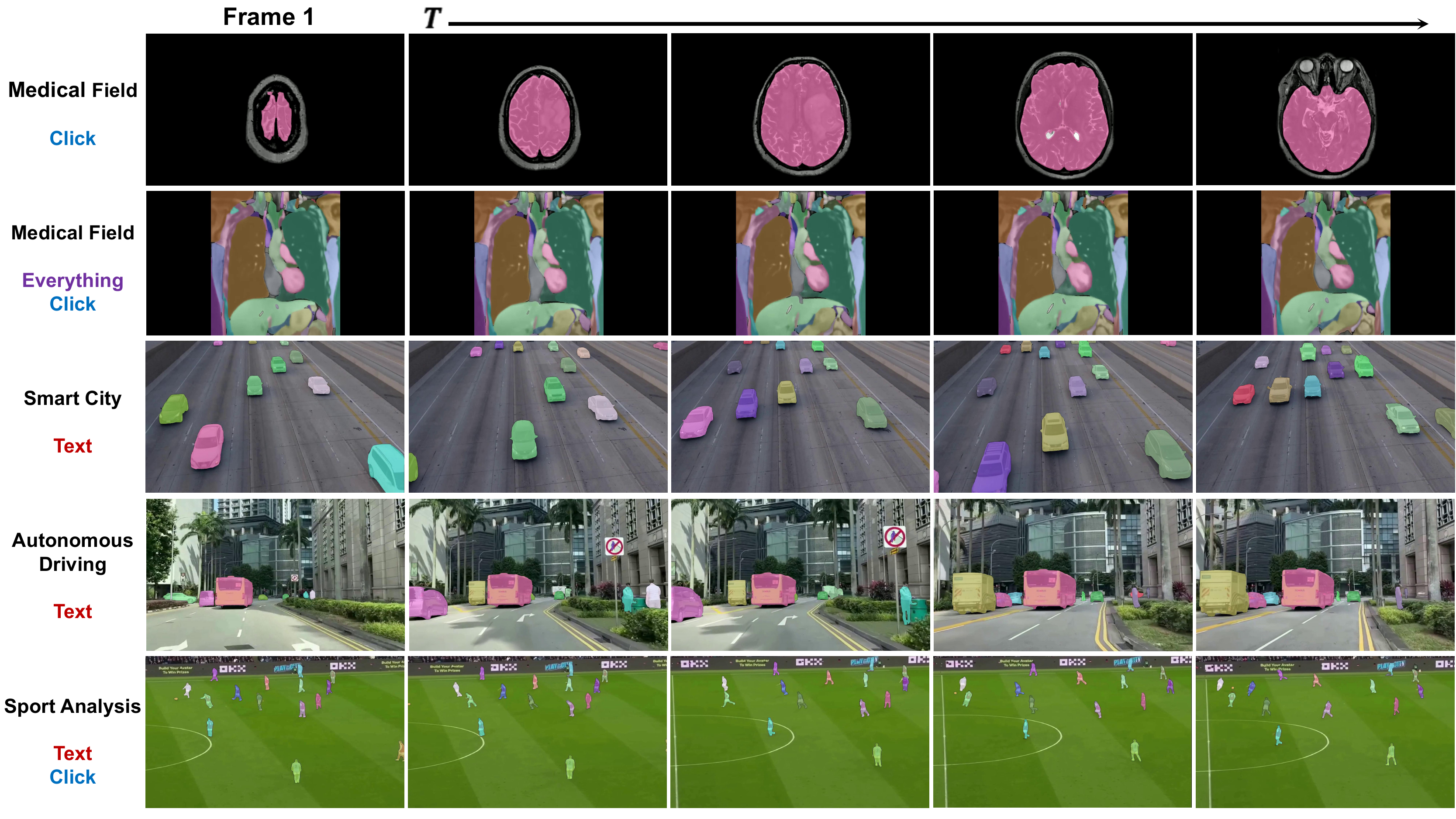}
\caption{The qualitative results of SAM-Track across different domains. We demonstrate the application of SAM-Track in various fields using interactive tracking, automatic tracking, and fusion tracking modes.}\label{fig:app}
\end{center}
\end{figure}

%% file: Sections/conclusion.tex
\section{Conclusion}\label{sec:conclusion}
This report proposes SAM-Track, a unified video segmentation model that supports multimodal interactions and is capable of tracking new objects of interest that appear midway through the video. With the efficient DeAOT tracking mode, SAM-Track can track multiple objects with fast inference speed. The efficiency and versatility of SAM-Track make it applicable to various fields with different requirements.We hope that SAM-Track can serve as a reliable baseline and accelerate the application of VOS models in real-world scenarios.